\documentclass{article}
\usepackage{spconf,amsmath,graphicx}
\usepackage{multirow}
\usepackage{color}

\title{Hierarchical Pronunciation Assessment with Multi-Aspect Attention}
\name{Heejin Do$^{\star}$ \qquad Yunsu Kim$^{\star \dagger}$ \qquad  Gary Geunbae Lee$^{\star \dagger}$}
\address{$^{\star}$Graduate School of AI, POSTECH \\
$^{\dagger}$Department of Computer Science and Engineering, POSTECH}

\begin{document}
%
\maketitle
\begin{abstract}
Automatic pronunciation assessment is a major component of a computer-assisted pronunciation training system. To provide in-depth feedback, scoring pronunciation at various levels of granularity such as phoneme, word, and utterance, with diverse aspects such as accuracy, fluency, and completeness, is essential. However, existing multi-aspect multi-granularity methods simultaneously predict all aspects at all granularity levels; therefore, they have difficulty in capturing the linguistic hierarchy of phoneme, word, and utterance. This limitation further leads to neglecting intimate cross-aspect relations at the same linguistic unit. In this paper, we propose a Hierarchical Pronunciation Assessment with Multi-aspect Attention (HiPAMA) model, which hierarchically represents the granularity levels to directly capture their linguistic structures and introduces multi-aspect attention that reflects associations across aspects at the same level to create more connotative representations. By obtaining relational information from both the granularity- and aspect-side, HiPAMA can take full advantage of multi-task learning. Remarkable improvements in the experimental results on the speachocean762 datasets demonstrate the robustness of HiPAMA, particularly in the difficult-to-assess aspects.
\end{abstract}
\section{Introduction}
\label{sec:intro}
Computer Assisted Pronunciation Training (CAPT) systems provide extensive feedback to non-native (L2) language learners about their pronunciation along with automatically assessed scores \cite{neri2002feedback, lin2020automatic}. Their fairness and objectivity have led to extensive studies on the automatic pronunciation assessment task, which is an indispensable component of CAPT.

Although most early pronunciation assessment systems only evaluated a single phoneme-level score \cite{witt2000phone, wang2012improved, shi2020context}, current studies have evaluated other aspects of word or utterance such as prosody, and especially stress, fluency, and intonation, but using separate models \cite{1415269, cucchiarini2000quantitative, li2017intonation}. Recently, there has been an attempt to score multiple aspects at more than one granularity level of pronunciation using a single model, named GOPT \cite{gong2022transformer}, emphasizing that different aspects of multi-granularity are indeed correlated and that jointly learning them leads to comprehensive representation. Their structure, which predicts all aspects at all granularity levels in parallel, has significantly contributed to the pronunciation assessment tasks.

However, phonemes, words, and utterances have strong linguistic dependencies \cite{lin2020automatic}, which may not be captured by a parallel structure. This also leads to a lack of consideration of the internal interactions across different aspects evaluated at the same granularity. Moreover, although GOPT outperforms previous systems in most aspects, the assessment of the utterance \textit{Completeness} is extremely inferior compared to other aspects. An accurate evaluation is equally required for all aspects, thus improving the scoring quality of poor-performance aspects is a persistent challenge in multi-task learning.

In this paper, we propose a novel model named Hierarchical Pronunciation Assessment with Multi-aspect Attention (HiPAMA) that assesses the phoneme-, word-, and utterance-level scores one level at a time, incrementally encoding the pronunciation. HiPAMA directly incorporates the syntactic structure of an utterance, which is made of words that comprise phonemes. Furthermore, we introduce a multi-aspect attention mechanism, which attends to other aspects at the same level to obtain more representative features considering that aspects are correlated in nature. Unlike the Transformer encoder that fully connects all nodes, our model modulizes each aspect-assessment task within each level.

We conduct experiments on the publicly available speechocean762 dataset \cite{zhang2021speechocean762}, which includes one phoneme-, three word-, and five utterance-level score labels. HiPAMA remarkably outperforms the robust baseline \cite{gong2022transformer} on all three granularity levels of assessment tasks, achieving state-of-the-art results on most of the aspects. In particular, the assessment for utterance \textit{Completeness} has been significantly enhanced. The results show that HiPAMA is a robust architecture for multi-output regression with an explicit output hierarchy.

\section{Related Work}
\label{sec:format}
Early works on automatic pronunciation assessment assessed a single aspect score only at the phoneme level \cite{witt2000phone, wang2012improved, shi2020context} or various aspect scores of word or utterance levels separately \cite{1415269, cucchiarini2000quantitative, li2017intonation}. Emphasizing that phonemes, words, and sentences are not independent of each other, a hierarchical network that outputs a single score at each granularity has been proposed \cite{lin2020automatic}. Agreeing with their intention, we also utilize the hierarchical structure; however, our approach differs from theirs in that we predict multiple aspects at each granularity level and use a different overall architecture.

Using goodness of pronunciation (GOP) \cite{witt2000phone} features and the Transformer \cite{vaswani2017attention} encoder, recently proposed GOPT model predicts multiple aspects at multi-granularity levels in parallel \cite{gong2022transformer}. Based on their model, additional use of augmented GOP features has also been proposed \cite{chao20223m}. However, simultaneous prediction for different granularity levels does not consider that the quality at a lower level influences that at a higher level. To directly incorporate linguistic hierarchy of pronunciation, we propose HiPAMA. Furthermore, unlike the use of the Transformer encoder network, which holistically shares the layers among all aspect-assessment tasks, our hierarchical architecture internally contains each module for different aspects, supporting connections between them.

\section{Model Description}
\begin{figure}[htp]
    \centering
    \includegraphics[width=8.5cm]{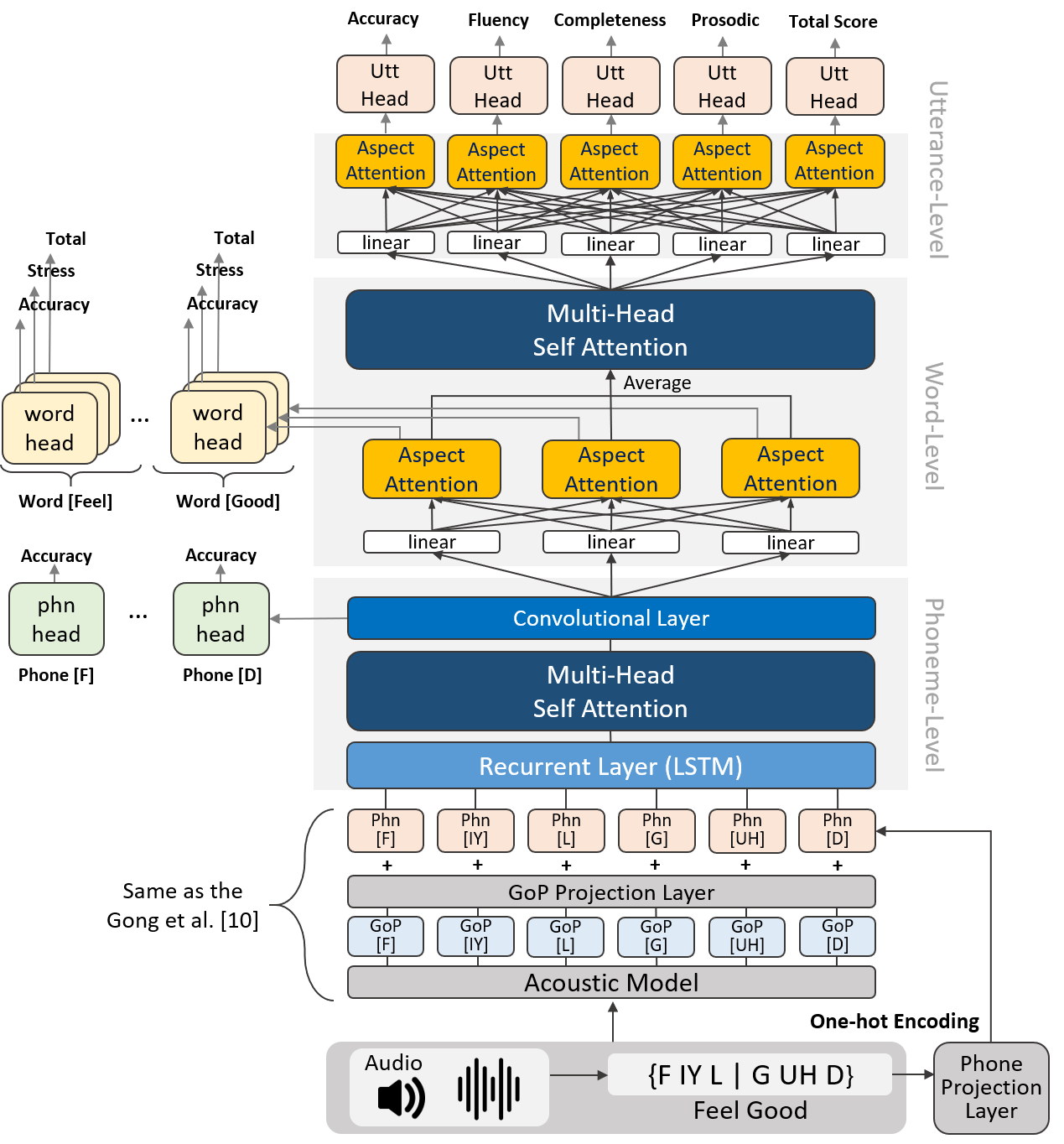}
    \caption{HiPAMA Architecture}
    \label{fig1}
\end{figure}

HiPAMA mainly includes three components at different granularity levels: phoneme, word, and utterance level. At each level, HiPAMA contains modulized layers for each aspect-assessment task (Figure~\ref{fig1}).

\subsection{Model Inputs}
For the input parts of the model, we follow the baseline model \cite{gong2022transformer} for a fair comparison. In particular, the automatic speech recognition (ASR) acoustic module takes the audio and its transcription as the input and predicts the sequence of frame-level phonetic posterior probabilities. These probabilities are transformed into 84-dimensional GOP features:
\begin{small}\begin{equation}
[LPP(p_1),\dots, LPP(p_{42}), LPR(p_1|p),\dots, LPR(p_{42}|p)]
\end{equation}\end{small}
where \begin{small}$LPP$\end{small} and \begin{small}$LPR$\end{small} denote the log phone posterior and log posterior ratio of a phone $p$, respectively. A phone $p$ is one of the pure phones in the acoustic model. Then, a dense layer projects the GOP features into 24 dimensions \cite{gong2022transformer}. Apart from GOP features, phoneme embedding is computed by projecting each phoneme-level one-hot encoding into 24 dimensions. Then, the GOP features and phoneme embedding are added and used as the input to our model.

\begin{table*}[t]
\centering
\scalebox{
0.85}{
\begin{tabular}{|l|cc|ccc|ccccc|}
\hline
& \multicolumn{2}{c|}{Phoneme Score} & \multicolumn{3}{c|}{Word Score (PCC)} & \multicolumn{5}{c|}{Utterance Score (PCC)} \\
\hline
\textbf{Model} & Acc(MSE ↓) & Acc(PCC ↑) & Acc ↑ & Stress ↑ & Total ↑ & Acc ↑ & Comp ↑ & Fluency ↑ & Prosody ↑ & Total ↑ \\
\hline
\multirow{2}{*}{LSTM} & 0.089 & 0.587 & 0.511 & 0.297 & 0.524 & 0.717 & 0.123 & 0.741 & 0.744 & 0.743\\
 & \small{±0.002} & \small{±0.014} & \small{±0.014} & \small{±0.012} & \small{±0.011} & \small{±0.004} & \small{±0.143} & \small{±0.010} & \small{±0.006} & \small{±0.006} \\
\cline{2-11}
\multirow{2}{*}{Gong et al.\cite{gong2022transformer}} & 0.085 & 0.612 & 0.533 & 0.291 & 0.549 & 0.714 & 0.155 & 0.753 & 0.760 & 0.742\\
 &  \small{±0.001} & \small{±0.003} & \small{±0.004} & \small{±0.030} & \small{±0.002} & \small{±0.004} & \small{±0.039} & \small{±0.008} & \small{±0.006} & \small{±0.005}\\
\cline{2-11}
\multirow{2}{*}{Gong et al.\cite{gong2022transformer}\small{-implement}} & 0.086 & 0.609 & 0.530 & 0.299 & 0.548 & 0.712 & 0.119 & \textbf{0.762} & \textbf{0.760} & 0.738\\
& \small{±0.001} & \small{±0.004} & \small{±0.003} & \small{±0.018} & \small{±0.003} & \small{±0.003} & \small{±0.264} & \small{±0.003} & \small{±0.008} & \small{±0.004}\\
\hline
\multirow{2}{*}{\textbf{HiPAMA}} & \textbf{0.084} & \textbf{0.616} & \textbf{0.575} & \textbf{0.320} & \textbf{0.591} &\textbf{0.730} & \textbf{0.276} & 0.749 & 0.751 & \textbf{0.754}\\
& \small{±0.001} & \small{±0.004} & \small{±0.004} & \small{±0.021} & \small{±0.004} & \small{±0.002} & \small{±0.177} & \small{±0.001} & \small{±0.002} & \small{±0.002}\\
\hline
\end{tabular}}
\caption{\label{tab1}
Experimental results with average MSE (phoneme level) and PCC (phoneme, word, and utterance level) scores and standard deviations of five different runs; \textbf{Acc} is \textit{Accuracy}; \textbf{Comp} is \textit{Completeness}; \begin{small}\textbf{-implement}\end{small} is implemented baseline model.}
\end{table*}

\subsection{Model Architecture}
Unlike the previous method that processes all granularity levels in parallel, the proposed method hierarchically represents each level of pronunciation. At each level, after obtaining the final representations for different aspects, a regression head for each aspect is followed to obtain the final aspect score.

First, for the phoneme-level representation, the long short-term memory (LSTM) \cite{hochreiter1997long} is applied to directly capture the sequential information of phonemes within the utterance \cite{li2017improving}. Then, a multi-head self-attention layer from \cite{vaswani2017attention} is applied to obtain the comprehensive contextual information for accurate assessment, which is then followed by a convolutional layer. The reason for utilizing a convolutional layer for phoneme-level representation is to better capture the local information corresponding to phonemes \cite{lee2009unsupervised, lee2016language}.

For word-level aspect scoring, encapsulated phoneme-level representations are passed to a separate module that assesses a different aspect. In the process of each module obtaining a representation for a specific aspect, our multi-aspect attention mechanism that computes attention scores with non-target aspect representations is introduced. This is motivated by the trait-attention mechanism applied to the automated essay scoring task \cite{ridley2021automated}, where each trait refers to the relevant information of other traits, and the term \textit{trait} is similar to the term \textit{aspect} in this task. We hypothesize that our multi-aspect attention method helps to better assess tricky aspects by accessing information about easy-to-assess aspects. In particular, first, for the $n$-th target aspect representation $\mathbf{a}^{n}$, non-target aspect representations $[\mathbf{a}^1,\dots,\mathbf{a}^{n-1},\mathbf{a}^{n+1},\dots,\mathbf{a}^N]$ are concatenated into $\mathbf{S}^n$. Then, a non-target matrix $\mathbf{A'}$ is obtained by applying attention pooling \cite{dong2017attention} to $\mathbf{S}^n$. For the cross-attention between the target aspect representation and non-target aspects, the following operation is performed:
\begin{small}\begin{eqnarray}
v_i^n&=&\frac{\mathrm{exp}(\mathrm{score}(\mathbf{a}^n, \mathbf{A'}_i))}{\sum_k \mathrm{exp}(\mathrm{score}(\mathbf{a}^n, \mathbf{A'}_k))}\label{eq12}\\
\mathbf{m}^n&=&\sum v_i^n \mathbf{A'}_{i} \label{eq13} \\
\mathbf{r}^n&=& \mathbf{a}^n + \mathbf{m}^n \label{eq13} 
\end{eqnarray}\end{small}
where the multi-aspect attention vector $\mathbf{m}^n$ is defined with the attention weight $v_i^n$. Then, by summing up the target aspect representation and the multi-aspect attention vector, the final n-th aspect representation $\mathbf{r}^n$ at specific level is obtained.

Finally, for the utterance-level aspect assessment, first, all word-level aspect representations are averaged. Then multi-head self-attention is applied with the aggregated word-level aspect representations, which can capture the contextual information between the words in the utterance. Then, as in the word level, the proposed multi-aspect attention mechanism is applied, so that each different utterance-level aspect benefits from other aspects, capturing relationships between them.

\subsection{Loss Function}
We use mean squared error (MSE) loss, which is commonly used for the pronunciation assessment task, as the loss function. The total loss is calculated as the sum of each granularity-level loss, each of which is an averaged loss of multiple aspects at that level:
\begin{small}\begin{equation}
\label{eq20}
    L_{total} = \sum_{m=1}^{M}\frac{1}{N}\sum_{n=1}^{N}L_{mn}
\end{equation}\end{small}
where $M$ and $N$ are the total numbers of granularity levels and aspects, respectively. $M$ is 3 when handling three levels.

\section{Experiments}
For the experiments, we use the publicly available speachocean762 dataset \cite{zhang2021speechocean762}, which is well designed for the multi-aspect multi-granularity pronunciation assessment task. It has score labels for three assessment granularity levels: utterance-, word-, and phoneme-level. For each level, there are multiple aspect scores: 1) \textit{Accuracy}, \textit{Completeness}, \textit{Fluency}, \textit{Prosody}, and \textit{Total} score for utterance-level; 2) \textit{Accuracy}, \textit{Stress}, and \textit{Total} score for word-level; and 3) \textit{Accuracy} score for phoneme-level. As in the baseline system \cite{gong2022transformer}, the utterance and word scores are scaled from (0-10) to (0-2) to achieve the same scale as the phoneme scores, and the same training and test set as the baseline system are used, each of which comprises 2500 utterances.

For a fair comparison, configurations other than the proposed parts are set to the same as the baseline, including the ASR acoustic model trained with Librispeech \cite{panayotov2015librispeech} 960-hour data. Specifically, HiPAMA is trained with the Adam optimizer, with a training epoch of 100, an initial learning rate of 1e-3, batch size of 25. Four heads are set for multi-head attention, and the dropout rate is set to 0.2 for utterance-level multi-head attention. Five runs with different random seeds are performed for all models, whose mean and standard deviation are reported. We use the Pearson correlation coefficient (PCC) as the evaluation metric along with MSE for the phoneme score as in \cite{gong2022transformer}.

\begin{table*}[t]
\centering
\scalebox{
0.83}{
\begin{tabular}{|l|cc|ccc|ccccc|}
\hline
& \multicolumn{2}{c|}{Phoneme Score} & \multicolumn{3}{c|}{Word Score (PCC)} & \multicolumn{5}{c|}{Utterance Score (PCC)} \\
\hline
Model & Acc(MSE ↓) & Acc(PCC ↑) & Acc ↑ & Stress ↑ & Total ↑ & Acc ↑ & Comp ↑ & Fluency ↑ & Prosody ↑ & Total ↑ \\
\hline
w/o Hi w/o MA & 0.085 & 0.609 & 0.566 & 0.299 & 0.582 & 0.723 & 0.112 & 0.740 & 0.744 & 0.746\\
w/o Hi \textbf{w/} MA & 0.085 & 0.614 & 0.571 & \textbf{0.323} & 0.586 & 0.723 & 0.238 & 0.741 & 0.743 & 0.746 \\
\hline
\textbf{w/} Hi \textbf{w/} MA (\textbf{HiPAMA}) & \textbf{0.084} & \textbf{0.616} & \textbf{0.575} & 0.320 & \textbf{0.591} &\textbf{0.730} & \textbf{0.276} & \textbf{0.749} & \textbf{0.751} & \textbf{0.754}\\
\hline
\end{tabular}}
\caption{\label{tab2}
Ablation results with average MSE (for phoneme level) and PCC (for phoneme, word, and utterance level) scores of five different runs. \textbf{Hi} and \textbf{MA} denote hierarchical structure and multi-aspect attention mechanism, respectively.}
\end{table*}

\section{Results and Discussion}
We compare HiPAMA with the previous state-of-the-art baseline model \cite{gong2022transformer}, which uses a 3-layer Transformer encoder, and the LSTM-based model, which was mainly compared with the baseline model (Table~\ref{tab1}). The results clearly demonstrate the strength of HiPAMA, which exhibit the highest PCC scores for most of the assessment tasks.

Assessment performance is remarkably improved for all aspects at all levels (average absolute improvement of 4.9\%), except \textit{Fluency} and \textit{Prosody} utterance scores (1.3\% and 0.9\% absolute decrease). These two aspects already had higher PCC values than other aspects, and HiPAMA shows comparable values with only a slight decrease. In particular, significant improvements are observed in the utterance \textit{Completeness} assessment task (131.9\% relative and 15.7\% absolute improvements), where previous models perform poorly.

\begin{figure}[t]
\begin{minipage}[b]{1.0\linewidth}
  \centering
  \centerline{\includegraphics[width=4.7cm]{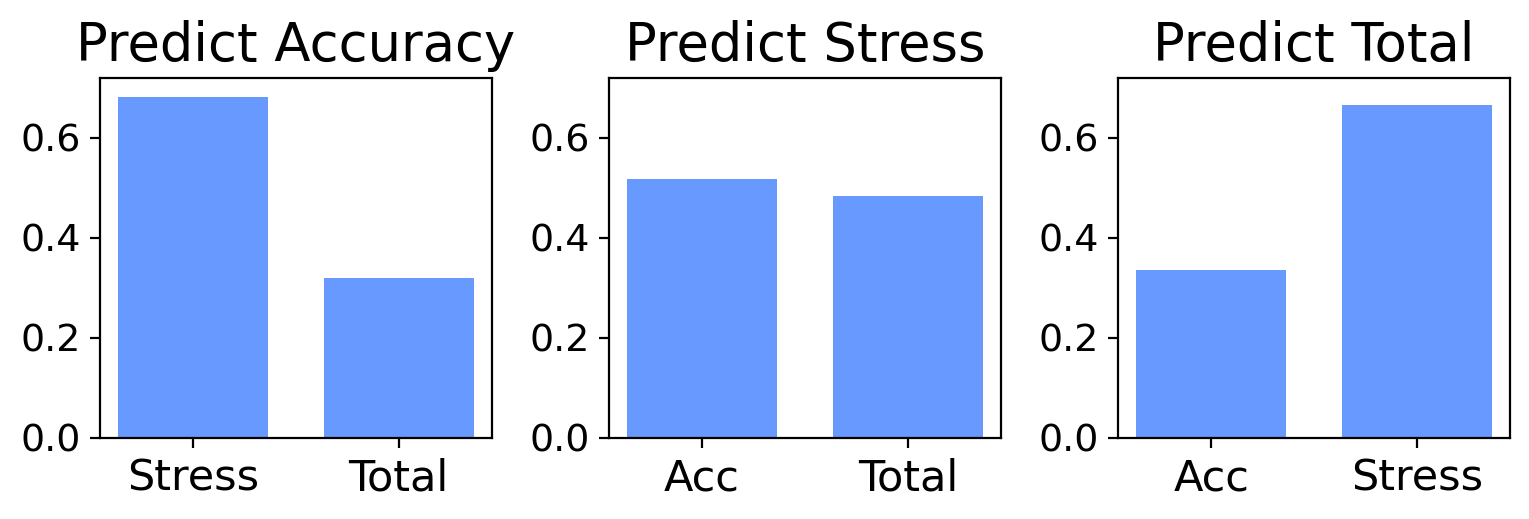}}
  \centerline{(a) Word level}\smallskip
\end{minipage}
\begin{minipage}[b]{1.0\linewidth}
  \centering
  \centerline{\includegraphics[width=8.6cm]{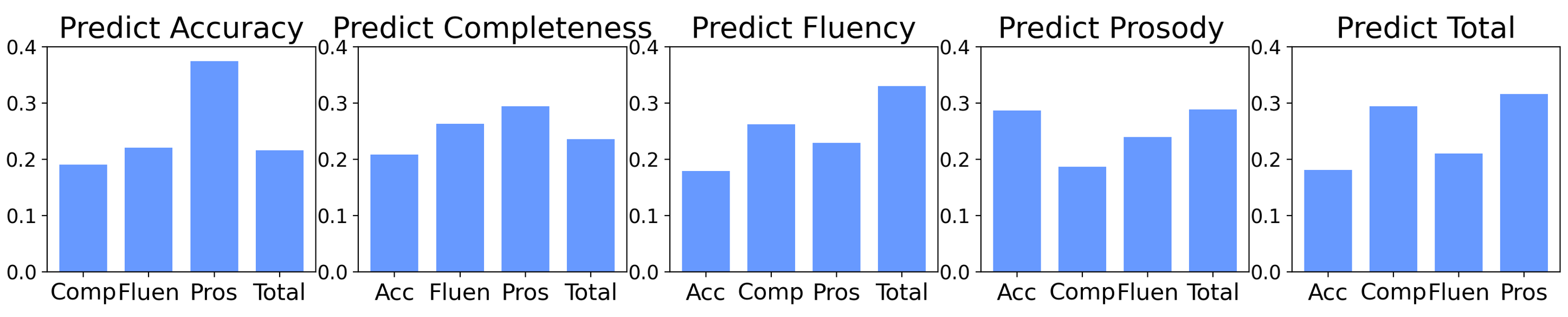}}
  \centerline{(b) Utterance Level}
\end{minipage}
\caption{Multi-aspect attention weights (run with seed 0) for other aspects when predicting each aspect score.}
\label{fig:attn}
\end{figure}

Further examination of multi-aspect attention weights enhances the explainability of our method, revealing other aspects that are being referenced when scoring a specific aspect (Figure~\ref{fig:attn}). For example, at the word level (a), the \textit{Accuracy} aspect highly focuses on the \textit{Stress} representation, while the utterance-level (b) \textit{Accuracy} aspect gives high weights to \textit{Prosody}. This result indicates that our method reflects the real-world assessment process because the scoring metric of word-level \textit{Accuracy} includes the criterion of accents \cite{zhang2021speechocean762}, and the accuracy of sentence-level pronunciation closely depends on the prosody \cite{sonia2016importance} in actual language learning. For the scoring of weak aspects, such as word \textit{Stress} and utterance \textit{Completeness}, the weight distribution is very consistent, showing that they are evenly influenced by all other aspects. These findings indicate that the proposed multi-aspect attention mechanism can assist weak aspect assessment by referring to other strong aspect representations. In the assessment of \textit{Total} score at each level, the model refers to the highly improved aspects, word \textit{Stress} and utterance \textit{Completeness}, a lot. This result implies that the performance improvements in relatively difficult assessment tasks contribute to overall scoring performance improvements (Table~\ref{tab1}). The improved assessment quality for one aspect further enhances the assessment task for the other aspect that highly attends to it, cooperatively influencing each other.

\subsection{Ablation Studies}
To investigate the effect of each component of HiPAMA, we conduct ablation studies in Table~\ref{tab2}. First, we experiment with a model without hierarchical structure and multi-aspect attention, which predicts all aspects of all granularity levels in parallel right after the LSTM, multi-head self-attention, and convolutional layer (1st row). Second, we evaluate a model in which only multi-aspect attention is added to the first model (2nd row). Note that HiPAMA is a model with a hierarchical structure applied to the second model. 

Because of the addition of multi-aspect attention, the performance improvements for previously difficult-to-assess aspects (word \textit{Stress} and utterance \textit{Completeness}) are considerably greater than other aspects. Furthermore, applying the hierarchical structure affords overall performance improvements; particularly in the last step, utterance level, the assessment performance significantly increased. The remarkable point is that jointly applying both of our approaches increases their synergistic impact for multiple aspect-assessment tasks.

\subsection{Effects of Model Size}
The baseline model uses a 3-layer Transformer encoder of 26.58k parameters, while HiPAMA has 31.64k parameters. To prevent a favorable condition for HiPAMA regarding the model size, we additionally compare HiPAMA with a larger baseline model: a 4-layer Transformer encoder-based model of 33.73k parameters (Table~\ref{tab4}). Although the baseline model itself slightly improves as the number of parameters increases, HiPAMA still outperforms both models even with a slightly smaller model size. These results prove that the robustness of HiPAMA is not due to the increased number of parameters.

\begin{table}[t]
\centering
\scalebox{
0.83}{
\begin{tabular}{|l|c|cc|c|c|}
\hline
&  & \multicolumn{2}{c|}{Phoneme} & Word & Utterance \\
\hline
Model & \# param & MSE ↓ & PCC ↑ & Avg PCC↑ & Avg PCC↑\\
\hline
3layer-base & 26.58k & 0.086 & 0.609 & 0.459 & 0.618 \\
4layer & 33.73k & 0.086 & 0.609 & 0.452 & 0.628 \\
\hline
\textbf{HiPAMA} & 31.64k & \textbf{0.084} & \textbf{0.616} & \textbf{0.495} & \textbf{0.652} \\
\hline
\end{tabular}}
\caption{\label{tab4}
Effect of the number of parameters (\# param). Because of the space limit, only the average score of the aspects has been reported for each granularity.}
\end{table}

\section{Conclusion}
In this study, we propose the HiPAMA model. Our hierarchical structure is designed to capture the linguistic structures of phonemes, words, and utterances, while our multi-aspect attention mechanism enables representative encoding of internal connections between aspects at a specific level. Experimental results on the speechocean762 dataset verify that the HiPAMA architecture is indeed effective by outperforming the previous state-of-the-art model.

\noindent \\ \textbf{Acknowledgements:}
This research was supported by the MSIT (Ministry of Science and ICT), Korea, under the ITRC (Information Technology Research Center) support program (IITP-2023-2020-0-01789) supervised by the IITP (Institute for Information \& Communications Technology Planning \& Evaluation) and supported by IITP grant funded by the Korea government (MSIT) (No.2022-0-00653, Voice Phishing Information Collection and Processing and Development of a Big Data Investigation Support System).


\bibliographystyle{IEEEbib}
\bibliography{strings,main}

\begin{thebibliography}{10}

\bibitem{neri2002feedback}
Ambra Neri, Catia Cucchiarini, and Helmer Strik,
\newblock ``Feedback in computer assisted pronunciation training: When
  technology meets pedagogy,''
\newblock 2002.

\bibitem{lin2020automatic}
Binghuai Lin, Liyuan Wang, Xiaoli Feng, and Jinsong Zhang,
\newblock ``Automatic scoring at multi-granularity for l2 pronunciation.,''
\newblock in {\em Interspeech}, 2020, pp. 3022--3026.

\bibitem{witt2000phone}
Silke~M Witt and Steve~J Young,
\newblock ``Phone-level pronunciation scoring and assessment for interactive
  language learning,''
\newblock {\em Speech communication}, vol. 30, no. 2-3, pp. 95--108, 2000.

\bibitem{wang2012improved}
Yow-Bang Wang and Lin-Shan Lee,
\newblock ``Improved approaches of modeling and detecting error patterns with
  empirical analysis for computer-aided pronunciation training,''
\newblock in {\em 2012 IEEE international conference on acoustics, speech and
  signal processing (ICASSP)}. IEEE, 2012, pp. 5049--5052.

\bibitem{shi2020context}
Jiatong Shi, Nan Huo, and Qin Jin,
\newblock ``Context-aware goodness of pronunciation for computer-assisted
  pronunciation training,''
\newblock {\em arXiv preprint arXiv:2008.08647}, 2020.

\bibitem{1415269}
J.~Tepperman and S.~Narayanan,
\newblock ``Automatic syllable stress detection using prosodic features for
  pronunciation evaluation of language learners,''
\newblock in {\em Proceedings. (ICASSP '05). IEEE International Conference on
  Acoustics, Speech, and Signal Processing, 2005.}, 2005, vol.~1, pp.
  I/937--I/940 Vol. 1.

\bibitem{cucchiarini2000quantitative}
Catia Cucchiarini, Helmer Strik, and Lou Boves,
\newblock ``Quantitative assessment of second language learners’ fluency by
  means of automatic speech recognition technology,''
\newblock {\em The Journal of the Acoustical Society of America}, vol. 107, no.
  2, pp. 989--999, 2000.

\bibitem{li2017intonation}
Kun Li, Xixin Wu, and Helen Meng,
\newblock ``Intonation classification for l2 english speech using
  multi-distribution deep neural networks,''
\newblock {\em Computer Speech \& Language}, vol. 43, pp. 18--33, 2017.

\bibitem{gong2022transformer}
Yuan Gong, Ziyi Chen, Iek-Heng Chu, Peng Chang, and James Glass,
\newblock ``Transformer-based multi-aspect multi-granularity non-native english
  speaker pronunciation assessment,''
\newblock in {\em ICASSP 2022-2022 IEEE International Conference on Acoustics,
  Speech and Signal Processing (ICASSP)}. IEEE, 2022, pp. 7262--7266.

\bibitem{zhang2021speechocean762}
Junbo Zhang, Zhiwen Zhang, Yongqing Wang, Zhiyong Yan, Qiong Song, Yukai Huang,
  Ke~Li, Daniel Povey, and Yujun Wang,
\newblock ``speechocean762: An open-source non-native english speech corpus for
  pronunciation assessment,''
\newblock {\em arXiv preprint arXiv:2104.01378}, 2021.

\bibitem{vaswani2017attention}
Ashish Vaswani, Noam Shazeer, Niki Parmar, Jakob Uszkoreit, Llion Jones,
  Aidan~N Gomez, {\L}ukasz Kaiser, and Illia Polosukhin,
\newblock ``Attention is all you need,''
\newblock {\em Advances in neural information processing systems}, vol. 30,
  2017.

\bibitem{chao20223m}
Fu-An Chao, Tien-Hong Lo, Tzu-I Wu, Yao-Ting Sung, and Berlin Chen,
\newblock ``3m: An effective multi-view, multi-granularity, and multi-aspect
  modeling approach to english pronunciation assessment,''
\newblock in {\em 2022 Asia-Pacific Signal and Information Processing
  Association Annual Summit and Conference (APSIPA ASC)}. IEEE, 2022, pp.
  575--582.

\bibitem{hochreiter1997long}
Sepp Hochreiter and J{\"u}rgen Schmidhuber,
\newblock ``Long short-term memory,''
\newblock {\em Neural computation}, vol. 9, no. 8, pp. 1735--1780, 1997.

\bibitem{li2017improving}
Wei Li, Nancy~F Chen, Sabato~Marco Siniscalchi, and Chin-Hui Lee,
\newblock ``Improving mispronunciation detection for non-native learners with
  multisource information and lstm-based deep models.,''
\newblock in {\em Interspeech}, 2017, pp. 2759--2763.

\bibitem{lee2009unsupervised}
Honglak Lee, Peter Pham, Yan Largman, and Andrew Ng,
\newblock ``Unsupervised feature learning for audio classification using
  convolutional deep belief networks,''
\newblock {\em Advances in neural information processing systems}, vol. 22,
  2009.

\bibitem{lee2016language}
Ann Lee et~al.,
\newblock {\em Language-independent methods for computer-assisted pronunciation
  training},
\newblock Ph.D. thesis, Massachusetts Institute of Technology, 2016.

\bibitem{ridley2021automated}
Robert Ridley, Liang He, Xin-yu Dai, Shujian Huang, and Jiajun Chen,
\newblock ``Automated cross-prompt scoring of essay traits,''
\newblock in {\em Proceedings of the AAAI conference on artificial
  intelligence}, 2021, vol.~35, pp. 13745--13753.

\bibitem{dong2017attention}
Fei Dong, Yue Zhang, and Jie Yang,
\newblock ``Attention-based recurrent convolutional neural network for
  automatic essay scoring,''
\newblock in {\em Proceedings of the 21st conference on computational natural
  language learning (CoNLL 2017)}, 2017, pp. 153--162.

\bibitem{panayotov2015librispeech}
Vassil Panayotov, Guoguo Chen, Daniel Povey, and Sanjeev Khudanpur,
\newblock ``Librispeech: an asr corpus based on public domain audio books,''
\newblock in {\em 2015 IEEE international conference on acoustics, speech and
  signal processing (ICASSP)}. IEEE, 2015, pp. 5206--5210.

\bibitem{sonia2016importance}
Belhoum Sonia and Benhattab Abdelkader~Lotfi,
\newblock ``The importance of prosody in a proper english pronunciation for efl
  learners,''
\newblock {\em Arab World English Journal (AWEJ) Volume}, vol. 7, 2016.

\end{thebibliography}

\end{document}